# Robust seed selection algorithm for k-means type algorithms

- **Optimal centroids using high density object**

-

[1]K. Karteeka Pavan, [2]Allam Appa Rao, [3]A.V. Dattatreya Rao, [4]G.R.Sridhar

[1]Department of Computer Applications, Rayapati Venkata Ranga Rao and Jagarlamudi Chadramouli College of Engineering, Guntur, India
[2]Jawaharlal Nehru Technological University, Kakinada, India
[3]Department of Statistics, Acharya Nagarjuna University, Guntur, India,
[4]Endocrine and Diabetes Centre, Andhra Pradesh, India
[1]kanadamkarteeka@gmail.com, [2]apparaoallam@gmail.com, [3]avdrao@gmail.com, [4]sridharvizag@gmail.com

**Abstract:** *Selection of initial seeds greatly affects the quality of the clusters and in k-means type algorithms. Most of the seed selection methods result different results in different independent runs. We propose a single, optimal, outlier insensitive seed selection algorithm for k-means type algorithms as extension to k-means++. The experimental results on synthetic, real and on microarray data sets demonstrated that effectiveness of the new algorithm in producing the clustering results*

## 1. Introduction

K-means is the most popular partitional clustering technique for its efficiency and simplicity in clustering large data sets (Forgy 1965; Macqueen 1967; Lloyd 1982, Wu et al., 2008). One of the major issues in the application K-Means-type algorithms in cluster analysis is, these are sensitive to the initial centroids or seeds. Therefore selecting a good set of initial seeds is very important. Many researchers introduce some methods to select good initial centers (Bradley and Fayyad, 1998; Deelers and Auwatanamongkol, 2007). Recently, Arthur and Vassilvitskii (2007) propose k-means++- a careful seeding for initial cluster centers to improve clustering results. Almost all algorithms produce different results in different independent runs. In this paper we propose a new seed selection algorithm, Single Pass Seed Selection (SPSS) that produces single, optimal solution which is outlier insensitive. The new algorithm is extension to k-means ++. K-means++ is a way of initializing k-means by choosing initial seeds with specific probabilities. The k-means++ selects first centroid and minimum probable distance that separates the centroids at random. Therefore different results are possible in different runs. For a good result the k-means++ has to be run number of times. The proposed SPSS algorithm selects the highest density point as the first centroid and also calculates minimum distance automatically using highest density point, , which is close to more number of other points in the data set. The objectives of the proposed SPSS algorithm are 1) to select optimal centroids 2) to generate single clustering solution instead most of the algorithms results different solutions in different independent runs. The quality of the clustering solution of SPSS is determined using various cluster validity measures and error rate is also identified using number of misclassifications. The experiments indicate that the SPSS algorithm converge k-means with unique solution and also it performs well on synthetic and real data sets.

## 2. Related Work

Inappropriate choice of number of clusters (Pham et al., 2004) and bad selection of initial seeds may yield poor results and may take more number of iterations to reach final solution. In this study we are concentrating on selection of initial seeds that greatly affect the quality of the clusters. One of the first schemes of centroids initialization was proposed by Ball and Hall (1967). Tou and Gonzales have proposed Simple Cluster Seeking (SCS) and is adopted in the FACTCLUS procedure. The SCS and the method suggested by Ball and Hall are sensitive to the parameter d and the presentation order of the inputs. Astrahan (1970) suggested using two distance parameters. The approach is very sensitive to the values of distance parameters and requires hierarchical clustering. Kaufman and Rousseeuw (1990) introduced a method that estimates the density through pair wise distance comparison and initializes the seed clusters using the input samples from the areas with high local density. A notable drawback of the method lies in its computational complexity. Given n input samples, at least $n(n-1)$ distance calculation are required. Katsavounidis et al. (1994) suggested a parameter less approach, which is called as the KKZ method based on the initials of all the authors. KKZ chooses the first centers near the "edge" of the data, by choosing the vector with the highest norm as the first center. Then, it chooses the next center to be the point that is farthest from the nearest seed in the set chosen so far. This method is very inexpensive $(O(kn))$ and is easy to implement. It does not depend on the order of points and is deterministic by nature as single run suffices to obtain the seeds. However, KKZ is sensitive to outliers, since it is selecting farthest point from the selected centroids. More recently, Arthur and Vassilvitskii (2007) proposed the k-means++ approach, which is similar to the KKZ (Katsavounidis et al., 1994) method. However, when choosing the seeds, they do not choose the farthest point from the already chosen seeds, but choose a point with a probability proportional to its distance from the already chosen seeds. In k-means++, the point will be chosen with the probability proportional to the minimum distance of this point from already chosen seeds. Note that due to the random selection of first seed and probabilistic selection of remaining seeds, different runs have to be performed to obtain a good clustering.

## 3.Methodology:

The proposed method, Single Pass Seed Selection (SPSS) algorithm is a modification to k-means++, is a method of initialization to k-means type algorithms. The SPSS initialize first seed and the minimum distance that separates the centroids based on highest density point, which is close to more number of other points in the data set. To show the modifications suggested in k-means++, k-means++ algorithm is presented here for ready reference.

**kmeans++ algorithm**

   k-means begins with an arbitrary set of cluster centers. k-means++ is a specific way of choosing these centers. The k-means++ is as follows:
Choose a set C of k initial centers from a point-set $(X_1, X_2,..,X_m)$:

   1. Choose one point uniformly at random from $(X_1,X_2,..,X_m)$ and add it to C
   2. For each point $X_i$, set $d(X_i)$ to be the distance between $X_i$ and the nearest point in C
   3. Choose a real number y uniformly at random between 0 and $d(X_1)^2+d(X_2)^2+...+d(X_m)^2$
   4. Find the unique integer i so that

5.  $d(X_1)^2+d(X_2)^2+...+d(X_i)^2 > = y > d(X_1)^2+d(X_2)^2+...+d(X_{(i-1)})^2$
6.  Add $X_i$ to C
7.  Repeat steps 2-5 until k centroids are found

## Single Pass Seed selection algorithm

Although the k-means++ is O (log k) competitive on all datasets, it also produce different clusters in different runs due to steps 1 and 3 in the algorithm. We propose a method for the steps 1, 3 of k-means++ to produce unique solution instead of different solutions, rather the proposed method-SPSS algorithm is a single pass algorithm:

Step 1:  Initialize the first centroid with a point which is close to more number of other points in the data set.

Step 3:  Assume that m (total number of points) points are distributed uniformly to k (number of clusters) clusters then each cluster is expected to contain m/k points. Compute the sum of the distances from the selected point (in step1) to first m/k nearest points and assume it as y.

## Steps of the algorithm

Choose a set C of k initial centers from a point-set $(X_1, X_2,.., X_m)$.  where k is number of clusters and m is the number of data points:

1.  Calculate distance matrix $Dist_{mxm}$ in which $dist(X_i,X_j)$ represents distance from $X_i$ to $X_j$.
2.  Find Sumv in which Sumv(i) is the sum of the distances from $X_i^{th}$ point to all other points.
3.  Find the index,h of minimum value of Sumv and find highest density point $X_h$ .
4.  Add $X_h$ to C as the first centroid.
5.  For each point $X_i$, set d $(X_i)$ to be  the  distance between $X_i$ and the nearest point in C.
6.  Find y as the sum of distances of first m/k nearest points from the $X_h$.
7.  Find the unique integr i so that
8.  $d(X_1)^2+d(X_2)^2+...+d(X_i)^2 > =  y>d(X_1)^2+d(X_2)^2+...+d(X_{(i-1)})^2$
9.  Add $X_i$ to C
10. Repeat steps 5-8 until k centroids are found

In k-means++ algorithm for selection of y, the number of passes in the worst case will be max$\{d(X_1)^2+d(X_2)^2+...+d(X_n)^2\}$ whereas it is equal to one in the proposed  SPSS algorithm. Therefore the SPSS is a single pass algorithm with unique solution while the k-means++ is not.

# 4. Experimental Results

The performance of SPSS is tested using both simulated and real data. The clustering results of SPSS is compared with k-means, k-means++ and fuzzy-k. These are implemented with the number of clusters as equal to the number of classes in the ground truth. The quality of the solutions of the algorithms is assessed with the Rand, Adjusted Rand, DB, CS and Silhouette cluster validity measures. The results of the proposed algorithm are also validated by determining the error rate. The error rate is defined as

$$err = \frac{N_{mis}}{m} * 100$$

, where $N_{mis}$ is the number of misclassifications and m is the number of elements of data set.

### 4.1 Experimental Data

The efficiency of SPSS is evaluated by conducting experiments on five artificial data sets, three real datasets down loaded from the web site UCI and two microarray data sets (two yeast data sets) downloaded from http://www.cs. washington.edu/homes/kayee/cluster (Yeung 2001).
The real data sets used:

1. Iris plants database (n = 150, d = 4, K = 3)
2. Glass (n = 214, d = 9, K = 6)
3. Wine (n = 178, d = 13, K = 3)

The real microarray data sets used:
The yeast cell cycle data (Cho *et al.*, 1998) showed the fluctuation of expression levels of approximately 6000 genes over two cell cycles (17 time points).

1. The first subset (the 5-phase criterion) consists of 384 genes whose expression levels peak at different time points corresponding to the five phases of cell cycle (Cho *et al.*, 1998).
2. The second subset (the MIPS criterion) consists of 237 genes corresponding to four categories in the MIPS database (Mewes *et al.*, 1999). The four categories (DNA synthesis and replication, organization of centrosome, nitrogen and sulphur metabolism, and ribosomal proteins) were shown to be reflected in clusters from the yeast cell cycle data (Tavazoie *et al.*, 1999).

The synthetic data sets:
The five synthetic data sets from $N_p(\mu, \sum)$ with specified mean vector and variance covariance matrix are as follows.

1. Number of elements, m=350, number of attributes, n=3, number of clusters, k =2 with

$$\mu 1 = \begin{pmatrix} 2 \\ 3 \\ 4 \end{pmatrix} \quad \Sigma 1 = \begin{pmatrix} 1 & 0.5 & 0.3333 \\ & 1 & 0.6667 \\ & & 1 \end{pmatrix} \quad \mu 2 = \begin{pmatrix} 7 \\ 6 \\ 9 \end{pmatrix} \quad \Sigma 2 = \begin{pmatrix} 1 & 1 & 1 \\ & 2 & 2 \\ & & 3 \end{pmatrix}$$

2. The data set with m=400, n=3, clusters=4 with

$$\mu 1 = \begin{pmatrix} -1 \\ -1 \end{pmatrix} \quad \Sigma 1 = \begin{pmatrix} 0.650 \\ & 0.65 \end{pmatrix} \quad \mu 2 = \begin{pmatrix} 2 \\ 2 \end{pmatrix} \quad \Sigma 2 = \begin{pmatrix} 1 & 0.7 \\ & 1 \end{pmatrix} \quad \mu 3 = \begin{pmatrix} -3 \\ +3 \end{pmatrix} \quad \Sigma 3 = \begin{pmatrix} 0.78 & 0 \\ & 0.78 \end{pmatrix} \quad \mu 4 = \begin{pmatrix} -6 \\ +4 \end{pmatrix} \quad \Sigma 4 = \begin{pmatrix} 0.5 & 0 \\ & 0.5 \end{pmatrix}$$

3. m=300, n=2, k=3;

$$\mu 1 = \begin{pmatrix} -1 \\ -1 \end{pmatrix} \quad \Sigma 1 = \begin{pmatrix} 1 & 0 \\ & 1 \end{pmatrix} \quad \mu 2 = \begin{pmatrix} 2 \\ 2 \end{pmatrix} \quad \Sigma 2 = \begin{pmatrix} 1 & 0 \\ & 1 \end{pmatrix} \quad \mu 3 = \begin{pmatrix} -3 \\ +3 \end{pmatrix} \quad \Sigma 3 = \begin{pmatrix} 0.7 & 0 \\ & 0.7 \end{pmatrix}$$

4. m=800, n=2, k=6;

$$\mu 1 = \begin{pmatrix} -1 \\ -1 \end{pmatrix} \quad \Sigma 1 = \begin{pmatrix} 0.650 \\ & 0.65 \end{pmatrix} \quad \mu 2 = \begin{pmatrix} -8 \\ -6 \end{pmatrix} \quad \Sigma 2 = \begin{pmatrix} 1 & 0.7 \\ & 1 \end{pmatrix} \quad \mu 3 = \begin{pmatrix} -3 \\ +6 \end{pmatrix} \quad \Sigma 3 = \begin{pmatrix} 0.2 & 0 \\ & 0.2 \end{pmatrix} \quad \mu 4 = \begin{pmatrix} -8 \\ +14 \end{pmatrix} \quad \Sigma 4 = \begin{pmatrix} 0.5 & 0 \\ & 0.5 \end{pmatrix}$$

$$\mu 5 = \begin{pmatrix} 10 \\ 12 \end{pmatrix} \quad \Sigma 5 = \begin{pmatrix} 0.3 & 0 \\ & 0.3 \end{pmatrix} \quad \mu 6 = \begin{pmatrix} +14 \\ -14 \end{pmatrix} \quad \Sigma 6 = \begin{pmatrix} 0.1 & 0 \\ & 0.1 \end{pmatrix}$$

5. m=180, n=8, k=3

$$\mu 1 = \begin{pmatrix} 1 \\ 1 \\ 2 \\ 1 \\ 0.5 \\ 2 \\ 1 \\ 0.5 \end{pmatrix} \quad \Sigma 1 = \begin{pmatrix} 1 & 0.5 & 0.333 & 0.25 & 0.2 & 0.1667 & 0.1429 & 0.125 \\ & 1 & 0.667 & 0.5 & 0.4 & 0.3333 & 0.2857 & 0.25 \\ & & 1 & 0.75 & 0.6 & 0.5 & 0.4286 & 0.375 \\ & & & 1 & 0.8 & 0.6667 & 0.5714 & 0.5 \\ & & & & 1 & 0.8333 & 0.7143 & 0.625 \\ & & & & & 1 & 0.8571 & 0.75 \\ & & & & & & 1 & 0.875 \\ & & & & & & & 1 \end{pmatrix} \quad \mu 2 = \begin{pmatrix} 1 \\ 1 \\ 1 \\ 1 \\ 1 \\ 1 \\ 1 \\ 1 \end{pmatrix}$$

$$\Sigma 2 = \begin{pmatrix} 1 & 1 & 1 & 1 & 1 & 1 & 1 & 1 \\ & 2 & 2 & 2 & 2 & 2 & 2 & 2 \\ & & 3 & 3 & 3 & 3 & 3 & 3 \\ & & & 4 & 4 & 4 & 4 & 4 \\ & & & & 5 & 5 & 5 & 5 \\ & & & & & 6 & 6 & 6 \\ & & & & & & 7 & 7 \\ & & & & & & & 8 \end{pmatrix} \quad \mu 3 = \begin{pmatrix} 1 \\ -2 \\ 0 \\ -1 \\ 0 \\ -1 \\ -2 \\ -2 \end{pmatrix} \quad \Sigma 3 = \begin{pmatrix} 1 & -1 & -1 & -1 & -1 & -1 & -1 & -1 \\ & 2 & 0 & 0 & 0 & 0 & 0 & 0 \\ & & 3 & 1 & 1 & 1 & 1 & 1 \\ & & & 4 & 2 & 2 & 2 & 2 \\ & & & & 5 & 3 & 3 & 3 \\ & & & & & 6 & 4 & 4 \\ & & & & & & 7 & 5 \\ & & & & & & & 8 \end{pmatrix}$$

## 4.2 Presentation of Results

Since the all existing algorithms produce different results in different individual runs, we have taken 40 independent runs of each algorithm. The performance of SPSS in comparison with k-means type algorithms is presented in terms of the validation measures Rand (Rand 1971), Adjusted Rand, DB (Davies and Bouldin 1979), CS (Chou et al. 2004) Silhouette metrics values and the error rate. The indices of various validation measures measure for different k-means type algorithms are calculated based on 40 independent runs of each algorithm and then taking the average, minimum and maximum values of the indices and error rate corresponding to the best possible algorithm along with corresponding index are compared with those obtained for SPSS algorithm and are tabulated in tables 1,2 and 3. The tables 1,2 and 3 respectively are meant for presenting the comparative results corresponding to average indices, error rate vs those of SPSS; minimum indices, error rate vs SPSS; and maximum vs SPSS. The minimum error rate that found and best performance values in 40 independent runs of each existing algorithm on each dataset is tabulated in Table2. The maximum error rate and the least performance values that found in 40 independent runs of each existing algorithm on each dataset is tabulated in Table3.

Table1. Comparison of SPSS with Mean values of 40 independent runs

| Dataset | Algorithm | k | Cluster Validity Measures | | | | | | Error rate in % |
| | | | CS | ARI | RI | HI | SIL | DB | erm |
|---|---|---|---|---|---|---|---|---|---|
| Synthetic1 | k-means | 2 | 0.645 | 0.92 | 0.96 | 0.92 | 0.839 | 0.467 | 0.236 |
| | k-means++ | | 0.567 | 0.925 | 0.962 | 0.925 | 0.839 | 0.466 | 1.914 |
| | fuzk | | 0.52 | 0.899 | 0.95 | 0.9 | 0.839 | 0.468 | 2.571 |
| | SPSS | | 0.725 | 0.932 | 0.966 | 0.932 | 0.839 | 0.465 | 1.714 |

| Synthetic2 | k-means | 4 | 1.178 | 0.821 | 0.927 | 0.854 | 0.718 | 0.58 | 19.1 |
| | k-means++ | | 1.21 | 0.883 | 0.953 | 0.907 | 0.776 | 0.519 | 7.16 |
| | fuzk | | 0.931 | 0.944 | 0.979 | 0.957 | 0.791 | 0.484 | 2.2 |
| | SPSS | | 0.812 | 0.939 | 0.977 | 0.953 | 0.792 | 0.527 | 2.4 |
| Synthetic3 | k-means | 3 | 0.87 | 0.957 | 0.98 | 0.96 | 0.813 | 0.509 | 2.242 |
| | k-means++ | | 0.92 | 0.97 | 0.987 | 0.974 | 0.823 | 0.761 | 1 |
| | fuzk | | 0.96 | 0.97 | 0.987 | 0.974 | 0.823 | 0.5 | 1 |
| | SPSS | | 0.657 | 0.97 | 0.987 | 0.974 | 0.823 | 0.507 | 1 |
| Synthetic4 | k-means | 6 | 0.72 | 0.816 | 0.941 | 0.882 | 0.82 | 0.407 | 51.27 |
| | k-means++ | | 0.62 | 0.958 | 0.988 | 0.976 | 0.932 | 0.222 | 10.96 |
| | fuzk | | 0.45 | 0.98 | 0.994 | 0.988 | 0.953 | 0.183 | 8.738 |
| | SPSS | | 0.723 | 1 | 1 | 1 | 0.975 | 0.144 | **0** |
| Synthetic5 | k-means | 3 | 1.78 | 0.197 | 0.62 | 0.24 | 0.396 | 1.176 | 53.9 |
| | k-means++ | | 1.678 | 0.201 | 0.622 | 0.244 | 0.398 | 1.133 | 54.42 |
| | fuzk | | 4.34 | 0.256 | 0.65 | 0.301 | 0.369 | 1.301 | 48.61 |
| | SPSS | | 1.821 | 0.183 | 0.614 | 0.228 | 0.392 | 1.279 | 52.22 |
| Iris | k-means | 3 | 0.607 | 0.774 | 0.892 | 0.785 | 0.804 | 0.463 | 15.77 |
| | k-means++ | | 0.712 | 0.796 | 0.904 | 0.807 | 0.804 | 0.461 | 13.37 |
| | fuzk | | 0.658 | 0.788 | 0.899 | 0.798 | 0.803 | 0.46 | 15.33 |
| | SPSS | | 1.962 | 0.44 | 0.72 | 0.441 | 0.799 | 0.582 | 50.67 |
| Wine | k-means | 3 | 0.612 | 0.295 | 0.675 | 0.35 | 0.694 | 0.569 | 34.58 |
| | k-means++ | | 0.678 | 0.305 | 0.681 | 0.362 | 0.694 | 0.562 | 33.54 |
| | fuzk | | 0.753 | 0.34 | 0.7 | 0.401 | 0.696 | 0.566 | 30.34 |
| | SPSS | | 0.813 | 0.337 | 0.699 | 0.398 | 0.696 | 0.601 | 30.34 |
| Glass | k-means | 6 | 0.967 | 0.245 | 0.691 | 0.382 | 0.507 | 0.901 | 55.86 |
| | k-means++ | | 1.523 | 0.259 | 0.683 | 0.365 | 0.548 | 0.871 | 56.1 |
| | fuzk | | 1.613 | 0.241 | 0.72 | 0.44 | 0.293 | 0.998 | 62.29 |
| | SPSS | | 1.512 | 0.252 | 0.722 | 0.444 | 0.382 | 1.061 | 45.79 |
| Yeast1 | k-means | 4 | 1.439 | 0.497 | 0.765 | 0.53 | 0.466 | 1.5 | 35.74 |
| | k-means++ | | 1.678 | 0.465 | 0.751 | 0.503 | 0.425 | 1.528 | 37.49 |
| | fuzk | | 1.679 | 0.43 | 0.734 | 0.468 | 0.37 | 2.012 | 39.18 |
| | SPSS | | 1.217 | 0.508 | 0.769 | 0.538 | 0.464 | 1.471 | 35.44 |
| Yeast2 | k-means | 5 | 1.721 | 0.447 | 0.803 | 0.607 | 0.438 | 1.307 | 38.35 |
| | k-means++ | | 1.521 | 0.436 | 0.801 | 0.603 | 0.421 | 1.292 | 40 |
| | fuzk | | 1.341 | 0.421 | 0.799 | 0.598 | 0.379 | 1.443 | 35.73 |
| | SPSS | | 2.567 | 0.456 | 0.804 | 0.608 | 0.453 | 1.236 | 43.23 |

Table2. Comparison of SPSS with best performance values of 40 independent runs

| Data Set | Algorithm | Validity Measure | | | | | | |
|---|---|---|---|---|---|---|---|---|
| | | ARI | RI | HI | SIL | DB | CS | miner |
| Synthetic 1 | K-means | 0.932 | 0.966 | 0.932 | 0.839 | 0.465 | 0.75 | 1.714 |
| | K-means++ | 0.932 | 0.966 | 0.932 | 0.839 | 0.465 | 0.75 | 1.714 |
| | Fuzzy-k | 0.899 | 0.95 | 0.9 | 0.839 | 0.468 | 0.732 | 2.571 |
| | SPSS | 0.932 | 0.966 | 0.034 | 0.839 | 0.465 | 0.725 | 1.714 |
| Synthetic 2 | K-means | 0.939 | 0.977 | 0.953 | 0.792 | 0.44 | 1.821 | 2.4 |
| | K-means++ | 0.939 | 0.977 | 0.953 | 0.792 | 0.44 | 1.805 | 2.4 |
| | Fuzzy-k | 0.944 | 0.979 | 0.957 | 0.791 | 0.44 | 0.931 | 2.2 |

| | | | | | | | |
|---|---|---|---|---|---|---|---|
| | SPSS | 0.939 | 0.977 | 0.023 | 0.792 | 0.527 | 0.812 | 2.4 |
| Synthetic 3 | K-means | 0.97 | 0.987 | 0.974 | 0.823 | 0.474 | 0.768 | 1 |
| | K-means++ | 0.97 | 0.987 | 0.974 | 0.823 | 0.474 | 1.701 | 1 |
| | Fuzzy-k | 0.97 | 0.987 | 0.974 | 0.823 | 0.474 | 0.749 | 1 |
| | SPSS | 0.97 | 0.987 | 0.013 | 0.823 | 0.507 | 0.657 | 1 |
| Synthetic 4 | K-means | 1 | 1 | 1 | 0.975 | 0.139 | 0.759 | 0 |
| | K-means++ | 1 | 1 | 1 | 0.975 | 0.142 | 0.403 | 0 |
| | Fuzzy-k | 1 | 1 | 1 | 0.975 | 0.127 | 0.412 | 0 |
| | SPSS | 1 | 1 | 0 | 0.975 | 0.144 | 0.723 | 0 |
| Synthetic 5 | K-means | 0.221 | 0.63 | 0.261 | 0.401 | 0.99 | 1.899 | 51.67 |
| | K-means++ | 0.221 | 0.63 | 0.261 | 0.401 | 0.99 | 1.899 | 51.67 |
| | Fuzzy-k | 0.261 | 0.653 | 0.305 | 0.371 | 1.054 | 4.83 | 46.67 |
| | SPSS | 0.183 | 0.614 | 0.386 | 0.392 | 1.279 | 1.821 | 52.22 |
| Iris | K-means | 0.886 | 0.95 | 0.899 | 0.806 | 0.411 | 0.753 | 4 |
| | K-means++ | 0.886 | 0.95 | 0.899 | 0.806 | 0.411 | 0.753 | 4 |
| | Fuzzy-k | 0.886 | 0.95 | 0.899 | 0.806 | 0.411 | 0.769 | 4 |
| | SPSS | 0.44 | 0.72 | 0.28 | 0.799 | 0.582 | 1.962 | 50.67 |
| Wine | K-means | 0.337 | 0.699 | 0.398 | 0.696 | 0.447 | 0.939 | 30.34 |
| | K-means++ | 0.337 | 0.699 | 0.398 | 0.696 | 0.447 | 0.939 | 30.34 |
| | Fuzzy-k | 0.347 | 0.704 | 0.408 | 0.696 | 0.488 | 0.929 | 29.78 |
| | SPSS | 0.337 | 0.699 | 0.301 | 0.696 | 0.601 | 0.813 | 30.34 |
| Glass | K-means | 0.287 | 0.728 | 0.456 | 0.656 | 0.744 | 1.917 | 44.86 |

| Data Set | Algorithm | ARI | RI | HI | SIL | DB | CS | maxer |
|---|---|---|---|---|---|---|---|---|
| | K-means++ | 0.288 | 0.725 | 0.45 | 0.729 | 0.522 | 1.745 | 46.73 |
| | Fuzzy-k | 0.263 | 0.733 | 0.467 | 0.317 | 0.883 | 3.995 | 48.13 |
| | SPSS | 0.252 | 0.722 | 0.278 | 0.382 | 1.061 | 1.512 | 45.79 |
| Yeast1 | K-means | 0.515 | 0.773 | 0.545 | 0.473 | 1.307 | 2.117 | 35.02 |
| | K-means++ | 0.515 | 0.772 | 0.545 | 0.473 | 1.376 | 2.006 | 35.02 |
| | Fuzzy-k | 0.453 | 0.744 | 0.489 | 0.396 | 1.722 | 16.91 | 37.55 |
| | SPSS | 0.508 | 0.769 | 0.231 | 0.464 | 1.471 | 2.012 | 35.44 |
| Yeast2 | K-means | 0.491 | 0.818 | 0.635 | 0.455 | 1.213 | 1.217 | 27.08 |
| | K-means++ | 0.497 | 0.82 | 0.64 | 0.514 | 1.092 | 1.666 | 26.3 |
| | Fuzzy-k | 0.478 | 0.812 | 0.625 | 0.43 | 1.296 | 6.384 | 27.86 |
| | SPSS | 0.456 | 0.804 | 0.196 | 0.453 | 1.236 | 2.567 | 43.23 |

Table3. Comparison of SPSS with least performance values of 40 independent runs

| Data Set | Algorithm | Validity Measure | | | | | | |
|---|---|---|---|---|---|---|---|---|
| | | ARI | RI | HI | SIL | DB | CS | maxer |
| Synthetic1 | K-means | 0.91 | 0.955 | 0.91 | 0.839 | 0.467 | 0.749 | 2.286 |
| | K-means++ | 0.91 | 0.955 | 0.91 | 0.839 | 0.467 | 0.749 | 2.286 |
| | Fuzzy-k | 0.899 | 0.95 | 0.9 | 0.839 | 0.468 | 0.732 | 2.571 |
| | SPSS | 0.932 | 0.966 | 0.034 | 0.839 | 0.465 | 0.725 | 1.714 |
| Synthetic2 | K-means | 0.561 | 0.82 | 0.641 | 0.503 | 0.904 | 0.936 | 67 |
| | K-means++ | 0.566 | 0.822 | 0.643 | 0.507 | 0.874 | 0.936 | 59.8 |
| | Fuzzy-k | 0.944 | 0.979 | 0.957 | 0.791 | 0.528 | 0.93 | 2.2 |
| | SPSS | 0.939 | 0.977 | 0.023 | 0.792 | 0.527 | 0.812 | 2.4 |
| Synthetic3 | K-means | 0.97 | 0.987 | 0.974 | 0.823 | 0.507 | 0.749 | 1 |
| | K-means++ | 0.432 | 0.718 | 0.436 | 0.44 | 0.962 | 0.749 | 50.67 |
| | Fuzzy-k | 0.97 | 0.987 | 0.974 | 0.823 | 0.507 | 0.731 | 1 |
| | SPSS | 0.97 | 0.987 | 0.013 | 0.823 | 0.507 | 0.657 | 1 |
| Synthetic4 | K-means | 0.574 | 0.851 | 0.703 | 0.504 | 0.727 | 0.233 | 0 |
| | K-means++ | 0.832 | 0.951 | 0.902 | 0.791 | 0.523 | 0.233 | 92.63 |
| | Fuzzy-k | 0.836 | 0.952 | 0.904 | 0.786 | 0.503 | 0.233 | 94.5 |
| | SPSS | 1 | 1 | 0 | 0.975 | 0.144 | 0.723 | 0 |
| Synthetic5 | K-means | 0.18 | 0.612 | 0.224 | 0.39 | 1.285 | 1.742 | 56.11 |
| | K-means++ | 0.18 | 0.612 | 0.224 | 0.39 | 1.285 | 1.742 | 56.11 |
| | Fuzzy-k | 0.23 | 0.636 | 0.272 | 0.317 | 1.477 | 2.137 | 48.89 |
| | SPSS | 0.183 | 0.614 | 0.386 | 0.392 | 1.279 | 1.821 | 52.22 |

| Iris | K-means | 0.44 | 0.72 | 0.441 | 0.798 | 0.582 | 0.607 | 51.33 |
|---|---|---|---|---|---|---|---|---|
| | K-means++ | 0.44 | 0.72 | 0.441 | 0.798 | 0.582 | 0.607 | 51.33 |
| | Fuzzy-k | 0.45 | 0.725 | 0.449 | 0.792 | 0.576 | 0.603 | 56 |
| | SPSS | 0.44 | 0.72 | 0.28 | 0.799 | 0.582 | 1.962 | 50.67 |
| Wine | K-means | 0.217 | 0.628 | 0.256 | 0.692 | 0.608 | 0.78 | 42.7 |
| | K-means++ | 0.217 | 0.628 | 0.256 | 0.687 | 0.608 | 0.774 | 42.7 |
| | Fuzzy-k | 0.332 | 0.696 | 0.392 | 0.695 | 0.601 | 0.914 | 30.9 |
| | SPSS | 0.337 | 0.699 | 0.301 | 0.696 | 0.601 | 0.813 | 30.34 |
| Glass | K-means | 0.152 | 0.666 | 0.333 | 0.207 | 1.168 | 0.966 | 67.29 |
| | K-means++ | 0.189 | 0.626 | 0.252 | 0.356 | 1.023 | 0.722 | 64.95 |
| | Fuzzy-k | 0.207 | 0.707 | 0.415 | 0.243 | 1.178 | 1.85 | 66.82 |
| | SPSS | 0.252 | 0.722 | 0.278 | 0.382 | 1.061 | 1.512 | 45.79 |
| Yeast1 | K-means | 0.246 | 0.658 | 0.315 | 0.184 | 1.757 | 1.509 | 80.17 |
| | K-means++ | 0.43 | 0.735 | 0.47 | 0.399 | 2.007 | 1.509 | 42.62 |
| | Fuzzy-k | 0.394 | 0.721 | 0.441 | 0.343 | 2.239 | 6.311 | 80.59 |
| | SPSS | 0.508 | 0.769 | 0.231 | 0.464 | 1.471 | 1.217 | 35.44 |
| Yeast2 | K-means | 0.361 | 0.784 | 0.568 | 0.339 | 1.489 | 1.53 | 57.03 |
| | K-means++ | 0.367 | 0.786 | 0.572 | 0.364 | 1.354 | 1.21 | 57.03 |
| | Fuzzy-k | 0.369 | 0.769 | 0.538 | 0.319 | 1.819 | 2.201 | 53.65 |
| | SPSS | 0.456 | 0.804 | 0.196 | 0.453 | 1.236 | 2.567 | 43.23 |

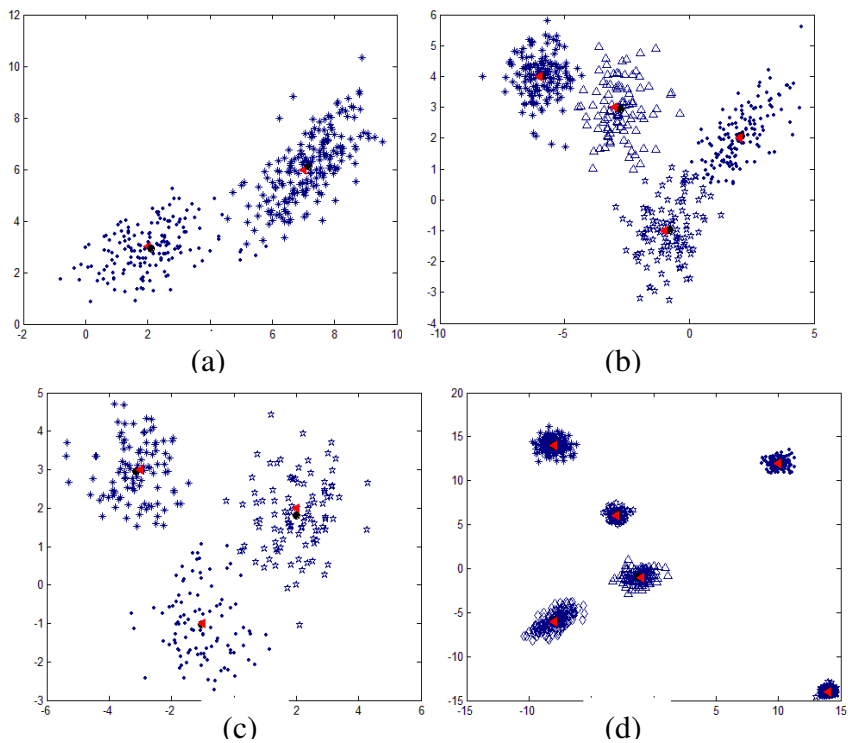

Figure 1. Clusters identified by SPSS, obtained centroids are marked with black circles and original centroids are marked with red triangles
(a)synthetic1 (b) synthetic2 (c) synthetic3 (d)synthetic4

Similarity matrix is a tool to judge a clustering visually. The clustering solutions of SPSS on the selected datasets have presented in the figure Figure2.

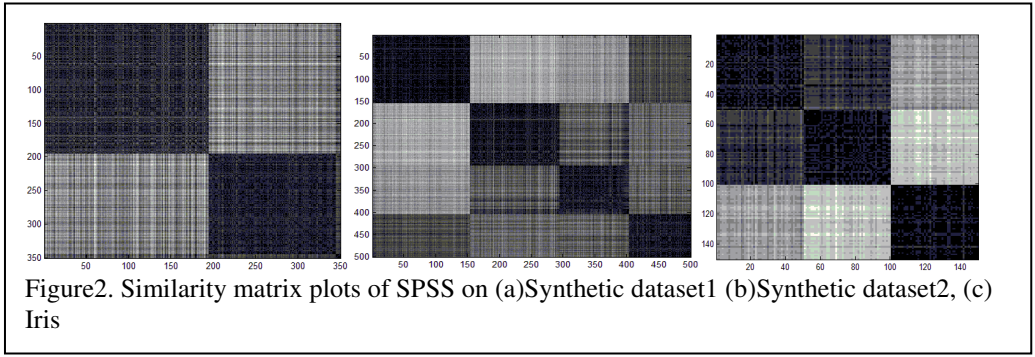

Figure2. Similarity matrix plots of SPSS on (a)Synthetic dataset1 (b)Synthetic dataset2, (c) Iris

Table4. SPSS performance in finding optimal Centroids

| Data set | Orignal Centroids | Obtained Centroids by SPSS |
|---|---|---|
| Synthetic1 | 7  6  9<br>2  3  4 | 7.1742    6.1103    9.2479<br>2.0975    2.9432    4.0614 |
| Synthetic2 | -6   4<br>2    2<br>-1   -1<br>-3   -3 | -5.9234    4.0052<br>2.0717    2.0794<br>-0.8052    -0.9848<br>-2.7743    2.9544 |
| Synthetic3 | -3   3<br>-1   -1<br>2   2 | -3.1467    2.9636<br>-1.0250    -1.0338<br>2.0025    1.8169 |
| Synthetic4 | -8  14<br>10   12<br>14   -14<br>-1   -1<br>-3   6<br>-8   -6 | -8.0344    14.0421<br>10.0285    12.0065<br>13.9763    -13.9768<br>-1.1876    -0.9205<br>-2.9580    6.0961<br>-7.9217    -5.9519 |

Figure3. Error rates of Glass dataset in 40 independent runs of kmeans, kmeans++ with SPSS

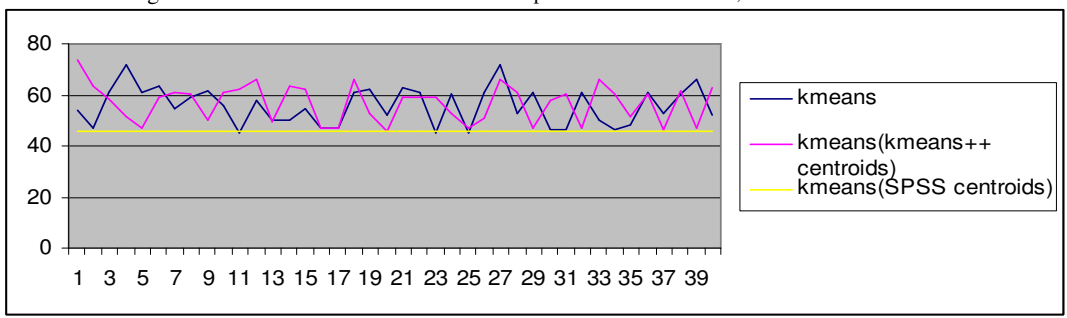

Figure4. Error rates of Yeast dataset in 40 independent runs of kmeans, kmeans++ with SPSS

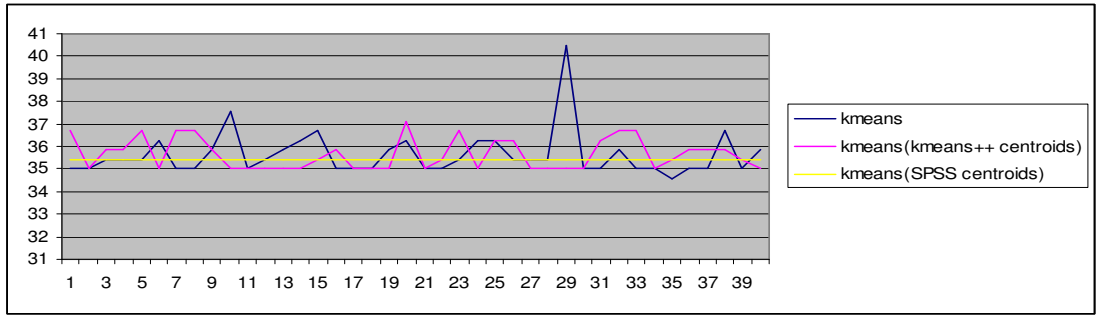

## Comments on the results of SPSS

- In the case of synthetic4 data set the SPSS outputs true clusters with error rate zero, and produced the value 1 for ARI,RI, HI indices where as the mean error rates are 51.67, 10.96 and 8.738 respectively for k-means, k-means++ and fuzzy-k.

- SPSS results 45.79 error rate for real data set Glass where as the k-means, k-means++, and fuzzy-k resulting the mean error rates as 55.86, 56.1, 62.29.

- For micro array data set yeast1, the existing algorithms are find clusters with mean error rates 37.49, 39.18, 35.44 but for the same the SPSS produced clusters with 35.44 error rate.

- In case of synthetic1, synthetic2, synthetic3, synthetic5, wine and yeast2 the SPSS equally performs with the means of other algorithms.

- In case of iris the SPSS resulting clusters with 50.67 error rate where others are determining with around 14 mean error rate.

- From the tables1 2, and 3 we can deduce that SPSS converged solution in least error rate which ever observed as the best value in 40 runs in the case of other algorithms. A comparison of SPSS with minimum error rates in 40 independent runs of other algorithms are summarized as follows.

Table 4 Comparative statement of error rates of existing algorithms with SPSS.

| Data set | Minimum error rates in 40 independent runs | SPSS Error rate |
|---|---|---|
| Synthetic1 | 1.714 | 1.714 |
| Synthetic2 | 2.2 | 2.4 |
| Synthetic3 | 1 | 1 |
| Synthetic4 | 0 | 0 |
| Synthetic5 | 51.67 | 52.2 |
| Wine | 29.8 | 30.3 |
| glass | 44.9 | 45.7 |
| Yeast1 | 35.2 | 35.4 |

- For iris the minimum error rate from the existing algorithms is 4, but the SPSS resulting as 50.67

- The error rate of SPSS for yeast2 is 43.23 where as the minimum error rate observed from the table 5.3.8 is 26.3.

- In the case of iris and yeast2 the SPSS is resulting poor clusters, but the error rates are not higher than the maximum error rates that found in 40 independent runs of other algorithms which are tabulated in the table3.
- The quality of SPSS is 82.58% in terms of Rand measure.
- The average improvement in terms of error rate over k-means is 10%, over k-means++ is 3% and over fuzzy-k is 2.8%. On an average there is nearly 5% improvement with robust solution in a single pass.

**Evaluation of Microarray data set clusters**

The clustering solutions of the microarray data sets of the proposed algorithm are visualized using the cluster profile plots (in parallel coordinates (Keim and Kriegel 1996) and the heatmap (Eisen *et al*. 1998) plots in the figures Figure 5 to Figure6. The expression profiles of coexpressed genes produce similar color patterns in heatmaps (Eisen *et al*. 1998). The similar color patterns of following figures Figure5 and 6 have demonstrated that the clustering solution of SPSS contain coexpressed gene groups i.e. the genes that are biologically similar.

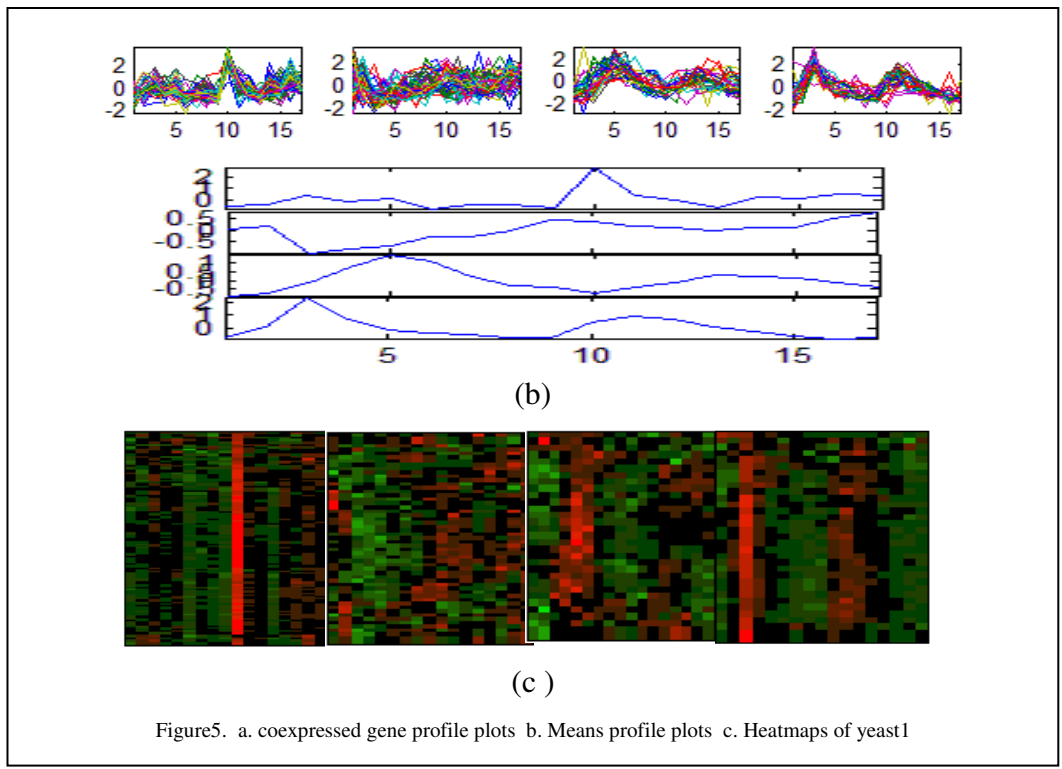

(b)

(c )

Figure5.  a. coexpressed gene profile plots  b. Means profile plots  c. Heatmaps of yeast1

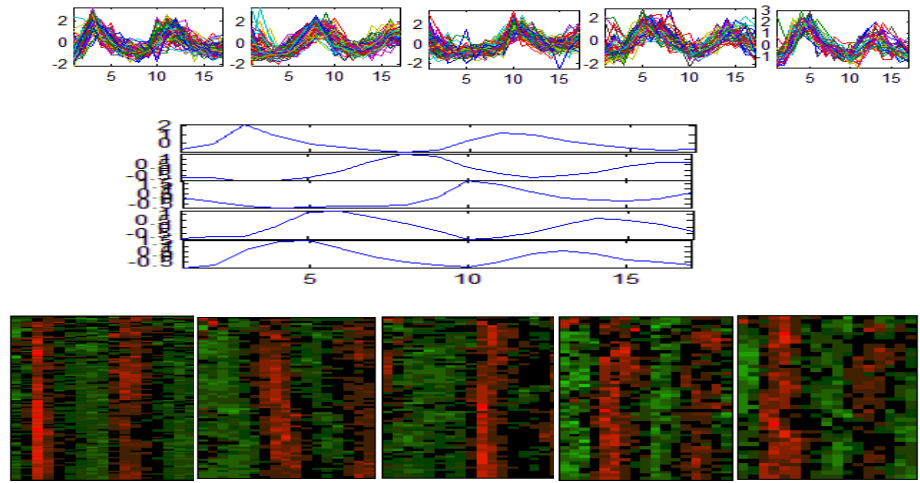

Figure6 a. Cluster profile plots b. mean profile plots c. heatmaps of yeast2

## Finding meaningful clusters from FatiGo

The evaluation criteria that help biologists to find the most valuable clusters from microarray data sets is different from the evaluation criteria in other clustering applications. The evaluation criteria of gene expressed profiles needs to be able to identify whether the genes in the same cluster have the similar biological function. We use Gene Ontology (GO) and P value to evaluate the clustering results. The FatiGO is a project (Al-Shahrour *et al.* (2004), http://www.fatigo.org), that determines the biological significance in terms of biological processes, molecular functions, and cellular components of the submitted clusters (Suresh *et al.* 2009). The results of the SPSS are submitted to the FatiGO to identify the gene enrichment of the clusters. The tool computes P value using hyper geometric distribution to determine the statistical significance of the association of a particular GO term with a group of genes in a cluster and evaluates whether the clusters have significant enrichment in one or more function groups. The P value is as follows:

$$P = 1 - \sum_{i=0}^{k} \frac{\binom{f}{i}\binom{g-f}{n-i}}{\binom{g}{n}}$$ where, n= total number of genes in a cluster, k = the number of

genes annotated to a specific GO term in a cluster, g=the number of genes in a whole genome, f= the number of genes annotated to a specific GO term in a whole genome and P is the probability of k genes annotated to a GO term among n genes of a cluster. P value is used to measure the gene enrichment of a microarray data cluster. If the majority of genes in a cluster biologically related, the P value of the category will be small. That is, the closer the P values to zero, the more the significance that the particular GO is term associated with the group of genes. We found that many P values are small, as shown in Figure 3 to Figure 5. Thus the proposed SPSS can find clusters with coexpressed genes. FatiGO produce the GO term for a given cluster of genes and a reference set of genes. The FatiGO computes various statistics for the given cluster of genes. It is observed that the percentage of genes in the cluster is considerably different from that of the reference cluster in almost all the functionalities. This implies that the correct genes are selected to remain in the same cluster. A sample of FatiGO results (GO terms) of a cluster of yeast1 as determined by SPSS is shown in the figures from Figure 7 to Figure 9, which are self explanatory.

Figure 7 Cluster2 molecular function by SPSS

Figure 8 Cluster2 cellular component by SPSS

| Term | Term size | Term size (in genome) | Term annotation % per list | | Annotated ids | Odds ratio (log+) | pvalue | Adjusted pvalue |
|---|---|---|---|---|---|---|---|---|
| mitotic cell cycle (GO:0000278) | 46 | 317 | list 1: 37.5%<br>list 2: 15.67% | | list 1: YAR007c,YBR088c...<br>list 2: YAR007c,YBR035c,YBR1... | 1.1723 | 0.006085 | 0.0285 |
| nucleobase, nucleoside, nucleotide and nucleic acid metabolic process (GO:0006139) | 108 | 1858 | list 1: 75%<br>list 2: 38.71% | | list 1: YAR007c,YBL035c...<br>list 2: YAR007c,YBL035c,YBR0... | 1.5581 | 0.0001889 | 0.001528 |
| DNA metabolic process (GO:0006259) | 83 | 712 | list 1: 68.75%<br>list 2: 28.11% | | list 1: YAR007c,YBL035c...<br>list 2: YAR007c,YBL035c,YBR0... | 1.7274 | 0.00001291 | 0.0001277 |
| DNA replication (GO:0006260) | 58 | 144 | list 1: 62.5%<br>list 2: 17.51% | | list 1: YAR007c,YBL035c...<br>list 2: YAR007c,YBL035c,YBR0... | 2.0606 | 3.206e-7 | 0.000004755 |
| DNA-dependent DNA replication (GO:0006261) | 45 | 84 | list 1: 43.68%<br>list 2: 13.82% | | list 1: YAR007c,YBL035c...<br>list 2: YAR007c,YBL035c,YBR0... | 1.7047 | 0.0000475 | 0.0004228 |
| RNA-dependent DNA replication (GO:0006278) | 16 | 25 | list 1: 21.88%<br>list 2: 4.15% | | list 1: YAR007c,YDL102w...<br>list 2: YAR007c,YDL102w,YJL1... | 1.8673 | 0.001507 | 0.008383 |
| DNA repair (GO:0006281) | 37 | 232 | list 1: 50%<br>list 2: 9.58% | | list 1: YAR007c,YBR087w...<br>list 2: YAR007c,YBR087w,YBR0... | 2.2336 | 2.815e-7 | 0.000004755 |
| nucleotide-excision repair (GO:0006289) | 18 | 44 | list 1: 25%<br>list 2: 4.61% | | list 1: YAR007c,YBR088c...<br>list 2: YAR007c,YBR088c,YDL1... | 1.9315 | 0.0005509 | 0.003772 |
| nitrogen compound metabolic process (GO:0006807) | 123 | 2106 | list 1: 78.12%<br>list 2: 45.18% | | list 1: YAR007c,YBL035c...<br>list 2: YAR007c,YBL035c,YBR0... | 1.4671 | 0.0005371 | 0.003772 |
| response to stress (GO:0006950) | 39 | 619 | list 1: 50%<br>list 2: 10.6% | | list 1: YAR007c,YBR087w...<br>list 2: YAR007c,YBR087w,YBR0... | 2.1324 | 7.126e-7 | 0.00000906 |

Figure 9 Biological process of Cluster 2 generated from SPSS

**Comments on the Microarray visualization results**

- The similar color patterns of heat maps of the microarray cluster profiles of SPSS have demonstrated that the expression profiles of the genes of a cluster are similar to each other.
- As determined by FatiGO (a web tool to evaluate micro array clusters using GO), the proposed algorithms increased the enrichment of genes of similar function within the cluster.
- The smaller P values (nearer to zero) in the FatiGO results indicates that the majority of genes in a cluster belong to one category and a particular GO term is associated with the group of genes

# Conclusion

k-means++ is a careful seeding for k-means. However, for good clustering results it has to repeat number of times and produces different results in different independent runs. The proposed SPSS algorithm is a single pass algorithm yielding unique solution with consistent clustering results compared to k-means++. Being the high density point is the first seed, the SPSS avoids different results that occur from random selection of initial seeds and the algorithm is insensitive to outliers in seed selection.

# REFERENCES

1. Al-Shahrour F, Dłaz-Uriarte R, and Dopazo J  2004  , "FatiGO: a web tool for finding significant associations of Gene Ontology terms with groups of genes", Bioinformatics, vol.20, pp.578–58.

2. Arthu, D. and S. Vassilvitskii, 2007. K-means++: The advantages of careful seeding. Proceeding of the 18[th] Annual ACM-SIAM Symposium of Discrete Analysis, Jan. 7-9, ACM Press, New Orleans, Louisiana, pp:1027-1035.

3. Astrahan, M.M., 1970. Speech analysis by clustering, or the Hyperphoneme method.

4. Ball, G.H. and D.J. Hall, 1967.  PROMENADE-an online pattern recognition system. Stanford Research Inst. Memo, Stanford University.

5. Berkhin, P., 2002. Survey of clustering data mining techniques. Technical Report, Accure Software, SanJose, CA.

6. Bradley, P.S. and U.M. Fayyad, 1998. Refining initial points for K-means clustering. Proceeding of the 15[th] International Conference on Machine Learning (ICML'98), July 24-27, ACM Press, Morgan Kaufmann, San Francisco, pp: 91-99.

7. ChoRJ Campbell MJ, Winzeler EA, Steinmetz L, Conway A, Wodicka L, Wolfsberg TG, Gabrielian AE, Landsman D, Lockhart DJ, and Davis RW 1998, "A genome-wide transcriptional analysis of the mitotic cell cycle," Mol. Cell, vol2, no.1, pp 65-73.

8. C Chou CH, Su MC, Lai E  2004  , "A new cluster validity measure and its application to image compression," Pattern Anal. Appl., vol. 7, no. 2, pp. 205–220

9. Davies DL and Bouldin DW,  1979  . A cluster separation measure. IEEE Transactions on Pattern Analysis and Machine Intelligence 1979, vol.1, pp.224-227

10. Deelers, S. and S. Auwatanamongkol, 2007. Enhancing K-means algorithm with initial cluster centers derived from data partitioning along the data axis with the highest variance. Proc. World Acad. Sci. Eng. Technol., 26:323-328.

11. Eisen, M.B., P.T. Spellman, P.O. Brown and D. Botstein, 1995.  Cluster analysis and display of genome-wide expression patterns.  Proc. Natl. Acad. Sci.USA, 95:14863-14868.

12. Fayyad, U.M., G. Piatetsky-Shapiro, P. Smyth and R. Uthurusamy, 1996. Advances in Knowledge Discovery and Data Mining.  AAAI/MIT  Press, ISBN: 0262560976, pp: 611.

13. Forgy E., Cluster analysis of multivariate data: Efficiency vs. interpretability of classifications, Biometrics 21, pp. 768, 1965.

14. Katsavounidis, I., C.C.J. Kuo and Z. Zhen, 1994. A new initialization  technique  for  generalized Lloyd iteration. IEEE.  Sig.  Process.  Lett., 1: 144-146. DOI: 10.1109/97.329844

15. Kaufman, L. and Rousseeuw, 1990. Finding Groups in Data: An Introduction to Cluster Analysis. Wiley,New York, SBN: 0471878766, pp: 342.

16. Keim DA and Kriegel HP 1996 , "Visualization techniques for mining large databases: a comparison", IEEE Transactions on Knowledge and Data Engineering vol.8 no.6, pp.923–938

17. Lloyd, S.P., 1982. Lease square quantization in PCM. IEEE Trans. Inform. Theor., 28: 129-136.

18. MacQueen, J.B., 1967. Some Method for Classification and Analysis of Multivariate Observations, Proceeding of the Berkeley Symposium on Mathematical Statistics and Probability, (MSP'67), Berkeley, University of California Press, pp: 281-297.

19. Mewes, H.W. , Heumann, K., Kaps, A. , Mayer, K. , Pfeiffer, F.stockerS., and Frishman,D 1999 . MIPS: a database for protein sequence and complete genomes. Nucleic Acids Research, 27:44-48.

20. Pham D. T., Dimov S. S., and Nguyen C. D.(2004), "Selection of k in K-means clustering," Mechanical Engineering Science, 219, pp. 103-119.

21. Rand WM 1971 , "Objective criteria for the evaluation of clustering methods. Journal of the American Statistical Association," vol.66, pp.846-850.

22. Rousseeuw J. and P. Silhouttes, 1987. A graphical aid to the interpretation and validation of cluster analysis.J. Comput. Applied Math., 20: 53-65.

23. Suresh K, Kundu K, Ghosh S, Das S and Abraham S 2009 "Data Clustering Using Multi-objective DE Algorithms" Fundamenta Informaticae, vol. XXI pp.1001–1024

24. Tavazoie S, Huges JD, Campbell MJ, Cho RJ and Church GM 1999 , "Systematic determination of genetic network architecture". Nature Genetics, vol.22, pp.281–285.

25. Wu, X., V. Kumar, J.R. Quinlan, J. Ghosh, D.J. Hand and D. Steinberg *et al*., 2008. Top10 algorithms in data mining. Knowl. Inform. Syst. J., 14: 1-37.